\documentclass[conference]{IEEEtran}
\IEEEoverridecommandlockouts
\usepackage{cite}
\usepackage{amsmath,amssymb,amsfonts}
\usepackage{algorithmic}
\usepackage{graphicx}
\usepackage{textcomp}
\usepackage{xcolor}
\usepackage{makecell}
\usepackage{changepage}
\usepackage{array}
\usepackage{placeins}
\usepackage{float}
\usepackage{multirow} % Required for \multirow command
\usepackage{threeparttable}
\def\BibTeX{{\rm B\kern-.05em{\sc i\kern-.025em b}\kern-.08em
    T\kern-.1667em\lower.7ex\hbox{E}\kern-.125emX}}

\begin{document}
% Generative AI-driven Immersive Simulation: A Knowledge-Aware Virtual Training Platform training platform for the for High Dose Rate (HDR) Brachytherapy in Vaginal Cancer Care
\title{Agentic AI-driven Immersive Simulation: A Knowledge-Aware Virtual Training Platform for High Dose Rate (HDR) Brachytherapy
}

% \author{Ronghua Xu}
% % \authornote{Both authors contributed equally to this research.}
% \email{ronghuax@mtu.edu}
% \author{Kepha Barasa}
% % \email{kwbarasa@mtu.edu}
% \author{Manoj Kumal}
% % \email{mkumal@mtu.edu}
% \author{Xinyun Liu}
% % \email{xinyunl@mtu.edu}
% \author{Weihua Zhou}
% % \email{whzhou@mtu.edu}

% \author{Xin Qian}
% % \authornote{Both authors contributed equally to this research.}
% \email{Xin.Qian@stonybrookmedicine.edu}
% \affiliation{%
%   \institution{Stony Brook University Hospital}
%   \city{Stony Brook}
%   \state{New York}

\author{\IEEEauthorblockN{Ronghua Xu${^{a*}}$, Kepha Barasa${^a}$, Manoj Kumal${^a}$, Xinyun Liu${^a}$, Weihua Zhou${^a}$, Xin Qian${^{b*}}$}
\IEEEauthorblockA{
$^{a}$Department of Applied Computing, Michigan Technological University, Houghton, MI 49931, USA\\
$^{b}$Stony Brook University Hospital, SUNY, Stony Brook, NY 11797, USA\\
$^{*}$Corresponding authors: ronghuax@mtu.edu, xin.qian@stonybrookmedicine.edu}
}

% \author{
% \IEEEauthorblockN{1\textsuperscript{st} Given Name Surname}
% \and
% \and
% \and
% \and
% \and

\maketitle

\begin{abstract}
The convergence of the Metaverse and Large Language Model (LLM)-based AI agent is catalyzing a shift toward autonomous, immersive, and personalized pedagogical frameworks in medical education. This paper presents a novel agentic AI-driven immersive simulation specifically designed for High Dose Rate (HDR) vaginal cylinder (VC) brachytherapy in cancer care.
% This system leverages virtual reality (VR) and mobile computing to create a highly realistic, safe virtual learning environment, where traineescan gain practical skills on HDR treatment without accessing physical female anatomy or directly handling actual radioactive sources.
By integrating Virtual Reality (VR) and mobile computing, the system establishes a high-fidelity, risk-free environment that allows trainees to master complex procedural skills without the facility or safety constraints posed by physical anatomy or live radioactive sources.
% In addition, we seamlessly integrate a knowledge-aware assistant system that uses retrieval argument generation (RAG) to incorporate HDR Brachytherapy clinical guidelines and an interactive agent to provide Natural Language Processing (NLP) interfaces and Hands-free guidance during complex medical training scenarios.
A core contribution of this work is the seamless integration of a knowledge-aware assistant leveraging Retrieval-Augmented Generation (RAG) to ground agent interactions in authoritative clinical guidelines.
This architecture also enables an interactive agent to provide natural language interfaces and hands-free, real-time guidance during intricate medical maneuvers.
% We implement the prototype consisting of a VR application running on Meta Quest 3 and a backend AI service on a local GPU workstation.
We validate the proposed system through a prototype deployment comprising a Meta Quest 3 interface linked to a local GPU-accelerated AI backend, demonstrating a feasible architecture for HDR brachytherapy simulation.
Experimental results indicate that the system maintains suitable end-to-end latency and high context precision, answer completeness, and relevance in the RAG-enhanced pedagogical support.
% Empirical evaluations verify the system’s efficacy, specifically in achieving low-latency query response times and in the high context precision, answer completeness and relevance of the RAG-enhanced pedagogical support.  

\end{abstract}


\section{Introduction}
\label{sec:introduction}
%% brief introduction to advancements in metaverse
% The advancements in Internet of Things (IoT) and communication technology, combined with fast-evolving artificial intelligence (AI)/machine learning (ML), enable ubiquitous services to improve human life.
% Over the past decade, technological advances and system integration have enabled an intelligent, connected world that significantly enhances human well-being and efficiency across domains such as digital healthcare, intelligent transportation, smart homes, and smart cities \cite{xu2024ar}.  
Through the combination of the prefix “meta” with the word “universe”, Metaverse aims for a hypothetical synthetic environment that seamlessly links the physical world and cyberspace.
From the technical aspect, the metaverse leverages key enabling technologies, like edge-cloud computing, Virtual Reality (VR), digital twins, and computer vision, to construct immersive virtual environments blending physical and digital spaces \cite{lee2024all}.
In recent years, the Metaverse, along with its core technologies, has demonstrated the potential to revolutionize multiple domains, such as healthcare, education, and smart industries \cite{xu2022full}.
%% -------------- evolution of LLM based agentic AI in healthcare -------------
At the same time, the rise of large models (LMs), usually referring to Large Language Models (LLMs) and multi-model vision-language models (VLMs), has led to the proliferation of LM-based agents that have promoted next-generation agentic AI to reshape global healthcare \cite{karunanayake2025next}.
% Traditional AI agents in healthcare rely on rule-based systems and static data-driven models for specific tasks with predefined objectives, such as classification and prediction.
% Agentic AI is transforming healthcare toward autonomous, adaptive, and goal-directed systems by leveraging key capabilities in perception, reasoning, memory, and action, along with human-in-the-loop (HITL) feedback \cite{wang2025large}.
LM-based agents adopt a sense-think-act operational architecture to integrate multimodal data sources, streamline clinic workflows, and support argument decision-making.
Thus, it promises to deliver continuous, responsive, longitudinal, and personalized healthcare service across diverse application domains, including radiation oncology, radiology, public health, and medical training \cite{banerjie2025agentic}.  

% Agentic AI healthcare survey: architecture \cite{banerjie2025agentic}

% Agentic AI healthcare survey including education \cite{xu2025comprehensive}

% Agentic AI hospital \cite{yao2025survey}

%% ========== narrow down to healthcare and medical training ====================
%% focuses on VR/AR and AI techniques
% The metaverse in nuclear medicine \cite{tang2024metaverse}.

% How AI supercharges VR app \cite{kuglerai}

% IAEA VR case study \cite{iaea_vr}.

\begin{figure}[t]
\centerline{\includegraphics[width=0.48\textwidth]{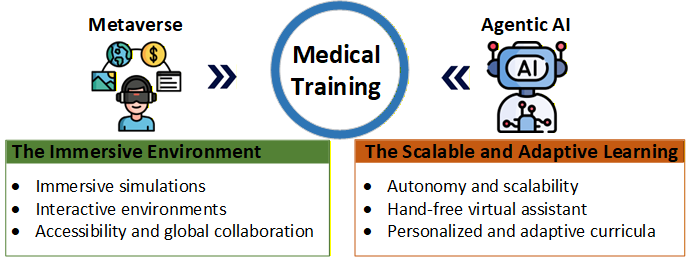}}
\caption{The integration of Metaverse and agentic AI in medical training}
\label{fig:VR_AI_healthcare}
\vspace{-15 pt}
\end{figure}

%% ========= Why VR+AI? How are they in organic synergy? ===========
The synergic integration of Metaverse and agentic AI promises to advance healthcare education, indicating a transformative shift toward more immersive, personalized, and efficient learning methods.
Figure \ref{fig:VR_AI_healthcare} illustrates the benefits of the metaverse and agentic AI in addressing challenges in medical education by creating immersive, scalable, and adaptive training systems.
As an enabling technology in the Metaverse, VR creates realistic, high-fidelity virtual scenarios (e.g., virtual patients) and clinical settings (e.g., operating rooms), allowing repeated practice of complex surgeries or hazardous conditions without patient risk.
In addition, gamified elements and interactive VR scenarios promote interactive environments that improve engagement and knowledge retention compared to traditional education based on passive learning.
Moreover, virtual platforms provide accessible educational services regardless of geographic location or time.
This facilitates global collaboration among learners and experts worldwide, enabling the sharing of learning experiences and case discussions.

%%------- challenges of the current HDR training system for medical school --------
High Dose Rate (HDR) vaginal cylinder (VC) brachytherapy is a critical yet highly complex procedure in radiation oncology, particularly for the treatment of vaginal and cervical cancers \cite{patankar2015high}.
Unlike external beam radiation, HDR brachytherapy requires precise internal placement of radioactive sources, demanding exceptional spatial reasoning, manual dexterity, and strict adherence to clinical protocols.
However, current training methodologies face several technical and safety challenges.
First, traditional HDR training requires a specific lab environment (e.g., shielding room, source security, and warning systems).
Even though the operation is conducted under strict guidelines, it inevitably poses safety risks to the patient and trainee by exposing them to actual radioactive sources.
% First, traditional training often involves practicing on physical manikins.
% This raises ethical concerns, even though the operation is under direct patient supervision.
% It also introduces risks to patient and trainee safety by handling actual radioactive sources.
In addition, training courses require qualified trainers (e.g., faculty or experts) to be continuously available onsite to provide guidance and feedback.
This dependency creates a bottleneck in the medical education pipeline, limiting the number of trainees due to facility and staffing resource constraints and making it difficult to scale standardized training across distributed geographic locations.
Moreover, trainees need to consult extensive clinical guidelines during complex procedures.
Manually consulting these resources during a procedure disrupts the ``flow'' of practice, increases cognitive load, and reduces achievement of learning objectives.

%--------justtify why VR+GenAI can help solve challenge in HDR training ----------
% Agentic AI can supercharge medical training by creating scalable, adaptive, and personalized learning environments.
% Agentic AI can act autonomously, make probabilistic decisions, manage complex workflows, and adapt the learning experience in real-time.
% AI agents can autonomously analyze learner actions in VR/AR and provide real-time, tailored feedback with limited human involvement.
% They can also support scalability by handling diverse data, distributed resources, and maintenance performance.
% Hand-free virtual assistants, such as real-time speech recognition, human-gesture tracking, and natural language processing, can significantly improve user-virtual environment interaction during training scenarios. 
% Furthermore, AI agents can monitor learners' performance, identify specific gaps, and create personalized, adaptive curricula that support individualized learning paths.

Several studies have shown the potential benefits of VR in medical education compared to traditional teaching methods, including VR models for cancer care training \cite{iaea_vr}, a VR assistant that improves learner skill development \cite{popovic2019simulation}, and VR-based informative ads to increase patient satisfaction \cite{chang2021virtual}.
Our solution offers several advantages over early work.
To address the aforementioned limitations in the current HDR brachytherapy training, this paper proposes an agentic AI-driven immersive simulation, an integrated framework that synthesizes VR and Agentic AI technologies to create a transformative pedagogical environment.
The main contributions of this paper are highlighted as follows: 

First, we use VR and mobile computing to develop a photorealistic, immersive learning environment for HDR VC brachytherapy.
This platform enables trainees to engage with anatomically accurate 3D models and virtualized radiopharmaceuticals, facilitating the mastery of HDR brachytherapy procedures.
By constructing a high-fidelity, risk-free virtual clinical training environment, our approach mitigates the privacy risks associated with physical pelvic examinations and eliminates the radiation safety hazards posed by direct handling of live radioactive sources.

In addition, our system uses a ``hands-free" virtual assistant powered by integrated real-time speech-to-text (STT) and neural text-to-speech (TTS) modules.
This framework facilitates seamless, bidirectional communication between the trainee and the virtual environment, enabling complex procedural inquiries in natural language.
By utilizing the VR hardware’s native audio system, we eliminate the need for manual controller input during intricate medical maneuvers, thereby reducing cognitive load and preserving the ergonomic fidelity of the clinical simulation.

Moreover, we propose and implement a domain-specific agent that uses Retrieval-Augmented Generation (RAG) to provide evidence-based guidance, thereby achieving knowledge-aware pedagogical intelligence.
By chunking and indexing authoritative HDR brachytherapy clinical guidelines into a high-dimensional vector database, our RAG agent generates responses strictly grounded in the medical literature.
Thus, trainees get real-time instructional support that is both clinically accurate and procedurally relevant to complex cancer care.
The proposed agentic AI-driven immersive simulation framework can be extended to other HDR procedures.
\textbf{The proposed framework complements the current physical HDR training system by providing unlimited, repeatable, and safe training environments that enable trainees to master complex procedures without requiring expensive, specialized facilities for pre-training practice.}

%% 5) Organization of this paper 
The remainder of this paper is organized as follows.
Section \ref{sec:background} describes the state-of-the-art on applications of VR and LLM-based agents in medical training.
Section \ref{sec:system} presents the system architecture consisting of a VR-based frontend and an RAG-enabled cognitive backend.
Section \ref{sec:experiment} presents numerical results that verify the performance of the proposed framework using a prototype implementation and expert-generated ground-truth datasets.
Section \ref{sec:conclusion} presents the conclusions and future direction of our study.

%%------------ need to reduce related work -------------
%% Kepha draft this section, R.Xu revise it
\section{Related Work}
\label{sec:background}
% This section introduces the existing solutions and relevant literature that are divided into two major categories: i) VR in medical training and ii) Agentic AI in healthcare education.
% The first category focuses on integrating VR/AR techniques into medical training procedures.
% The second category focuses on applying LM-based AI agent systems to healthcare education and training.  

%% Review existing works on VR/AR and even the metaverse in the medical industry
\subsection{Virtual Reality in Medical Training}
% The healthcare industry is undergoing rapid changes in medical training and education due to the emergence of digital technologies, including virtual reality (VR) and augmented reality (AR).
% The use of VR platforms in medical training environments enables future and current medical practitioners to experience realistic clinical scenarios, practice procedures in simulated settings, and receive immediate feedback, addressing major limitations of traditional pedagogy such as restricted access to cases, risks associated with live practice and the cost of operating sophisticated medical equipment \cite{chengoden_metaverse_2023}.
% The introduction of digital twin (DT) technology as a framework used to mirror physical entities in the real world to a virtual environment \cite{aloqaily_integrating_2023} has made a significant impact in synchronous medical education, overcoming barriers such as location and availability of medical equipment in resource-limited areas.

VR has been used in a wide range of medical education, planning, and diagnosis, including surgeries, imaging, especially radiology, and analysis of the human body at a cellular level \cite{chengoden2023metaverse}.
For instance, in 2020, neurosurgeons from John Hopkins Hospital performed operative planning surgery using AR headsets \cite{chengoden2023metaverse}.
The procedure utilized a transparent ocular display that projected anatomical images of the patients, providing an X-ray-like view to assist in fusing six spinal vertebrae and alleviating persistent back pain \cite{chengoden2023metaverse}.
% In this case, surgeons can explore multiple possibilities before performing the surgery on the actual surgery on the patient.
In ophthalmology, aspects of VR, such as improved visual and sensory feedback, are used in telesurgery, where a surgeon can perform a procedure from a remote location \cite{tan2022metaverse}.
Successful ophthalmology studies using telesurgery have been reported in animals, such as the repair of corneal lacerations in rabbit eyes \cite{tan2022metaverse}.
Compared to traditional catheterization laboratory mentor-based training,  
VR-enabled simulators \cite{popovic2019simulation} provide a consistent, standardized learning framework, enhancing trainee learning effectiveness through deliberate practice without risk to patients.
Extensive research indicates that integrating VR interventions significantly enhances the efficacy of patient education by improving health literacy and procedural comprehension \cite{chang2021virtual}.
% VR models for cancer treatment \cite{iaea_vr}, a VR assistant that improves learner skill development \cite{popovic2019simulation}, and VR-based informative ads to increase patient satisfaction \cite{chang2021virtual}.
To provide cost-effective solutions for cancer care training, the IAEA has developed a VR prototype \cite{iaea_vr} that helps oncologists, radiation therapists, and medical physicists become familiar with radiotherapy cancer set-ups and procedures.
Despite significant strides in developing immersive VR simulations for medical education, existing frameworks often serve as static or scripted environments that lack dynamic adaptation and cognitive intelligence.

\subsection{LLM-based Agents in Medical Education}
% With the accelerated development of artificial intelligence (AI), large language models (LLMs) and LLM-based agents have emerged as transformative tools in medical and healthcare applications \cite{chen_evaluating_2025} supporting tasks such as diagnosis and treatment assistance, medical education and training, scientific writing, and patient communication \cite{meng_application_2024}. 
In medical education, LLM-based agents offer a controlled, safe, and adaptable environment for patient simulation \cite{yao_survey_2025}.
For example, the proposed multi-agent framework, Evopatient \cite{du2025llms}, models a realistic diagnostic procedure across multiple phases of medical training.
EvoPatient integrates patient and doctor agents engaged in multi-turn dialogues, employing RAG \cite{lewis2020retrieval} to generate patient responses informed by historical medical data and personality traits derived from the Big Five taxonomy.
A key advancement of this system is its unsupervised co-evolution mechanism, which iteratively builds attention and trajectory libraries and refines conversation examples over time. This process gradually develops patient agents into standardized simulated patients while enabling doctor agents to improve questioning strategies. 
Similarly, the framework MEDCO \cite{wei_medco_2024}, a multi-agent co-pilot system, recreates realistic medical training settings that include patients, medical students, radiologists, and experts, thereby enhancing diagnostic reasoning and clinical communication \cite{yao_survey_2025}.
Furthermore, contemporary frameworks, such as the AI-Structured Clinical Examination (AI-SCE), prioritize the fundamental tenets of interpretability and explainability within diagnostic reasoning. By establishing rigorous transparency in how AI agents arrive at clinical conclusions, these systems facilitate the development of trustworthy LLM-based architectures suitable for the highly regulated nature of medical education \cite{yao2024medqa}.
% Furthermore, recent systems such as the AI Structured Clinical Examination (AI-SCE) emphasize interpretability and explainability in diagnostic reasoning, contributing to the development of trustworthy LLM-based agents for regulated clinical training environments \cite{yao2024medqa}.

% Despite the rapid advancement of generative AI in medical training, several persistent challenges continue to derail its effective integration into educational practice. These challenges include concerns about information accuracy, ethical responsibility, data privacy, and the limited capacity of current multi-agent systems to demonstrate emotional intelligence during patient interactions \cite{zhui2024impact}. Addressing these concerns is essential to ensure that AI-driven training environments not only uphold medical integrity and patient safety but also emulate the empathy and human sensitivity integral to effective clinical practice.

% %%------------ This section system design from the system level ------------
% %% R.Xu and Kepha
% %% 1) Core components of the system: VR application (Frontend) and AI service (Backend)
% %% 2) explain data and operation flows
% %% 3) How does the integrated system satisfy design objectives
\section{System Design}
\label{sec:system}
\begin{figure}[t]
\centerline{\includegraphics[width=0.48\textwidth]{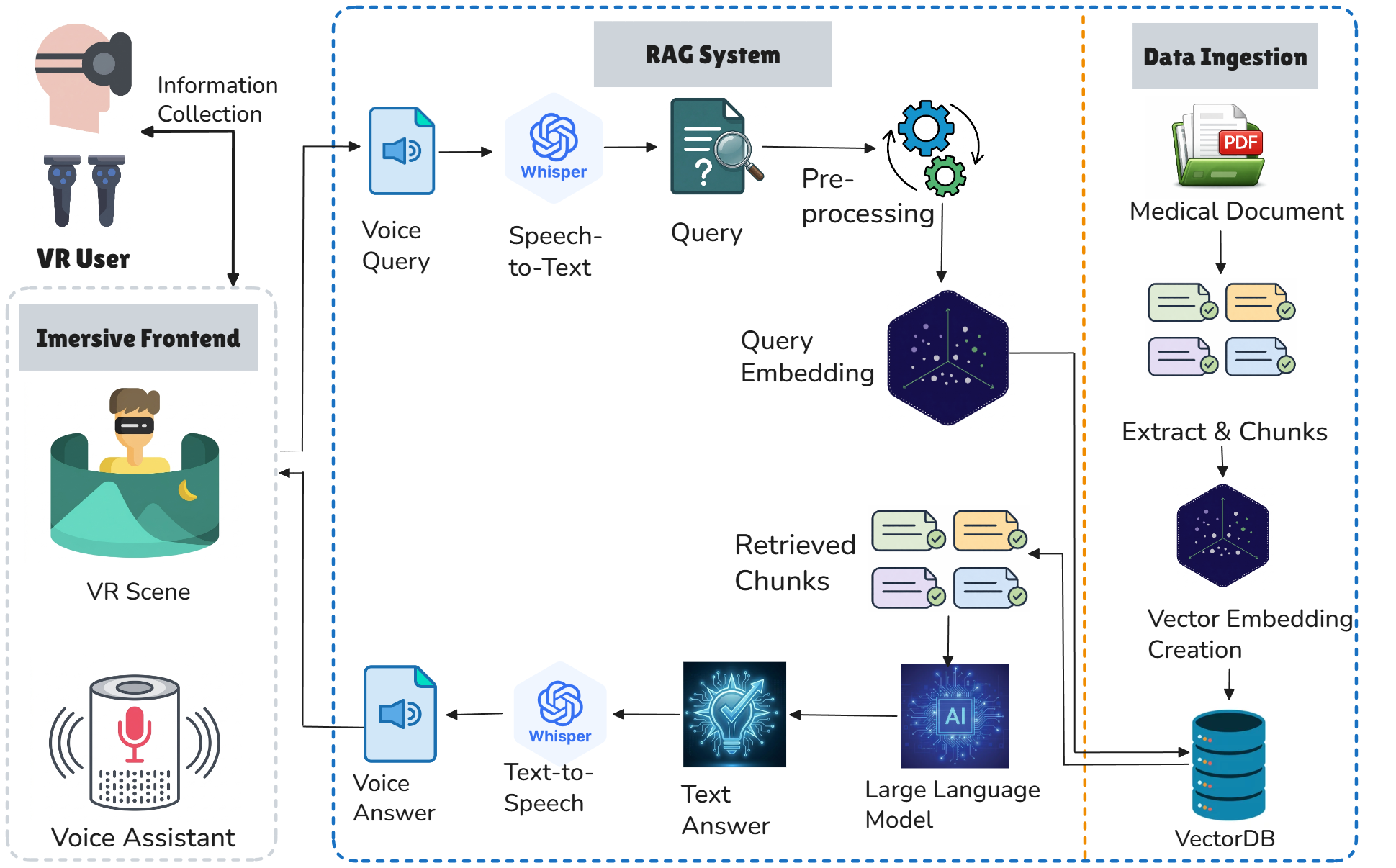}}
\caption{The system architecture of agentic AI-driven immersive simulation}
\label{fig:hdr_rag_platform}
\vspace{-10 pt}
\end{figure}

The proposed knowledge-aware virtual training platform is designed as a modular, distributed system based on an edge-cloud computing paradigm.
Figure \ref{fig:hdr_rag_platform} shows the architecture comprising an immersive frontend at a VR headset and a cognitive backend deployed on a high-performance server.
This architecture enables real-time, hands-free pedagogical support by offloading computationally intensive tasks (LLM inference and vector retrieval) to a dedicated server while maintaining high-fidelity visualization on the mobile VR headset.

\subsection{The Immersive Frontend}
The immersive frontend serves as the primary interaction layer for trainees through VR simulation and multimodal interface.
The VR simulation platform features a high-fidelity digital twin of an HDR brachytherapy suite, integrating high-resolution 3D clinical layouts with anatomically precise patient models and procedural instrumentation, including catheters, HDR afterloaders, and cylindrical applicators.
By prioritizing procedural asset fidelity, the system enables trainees to meticulously visualize the spatial orientation of the applicator relative to adjacent organs at risk (OARs), such as the bladder and rectum.
Such an immersive visualization is critical for mastering the nuances of optimal dose distribution.
This allows medical students to experiment with various applicator geometries to ensure precise, efficient, and safe radiation delivery while minimizing off-target exposure.
% A high-fidelity digital replication of an HDR Brachytherapy suite is created as VR scenes that show high-resolution 3D clinical layouts and models of patient anatomy and instruments (e.g., catheters, HDR afterloader, and cylinder applicator).
% Thanks to procedural asset fidelity, trainees can visualize the spatial relationship between the applicator and critical organs at risk.
% This allows for precise, fast, and safe radiation delivery, with various sizes available for optimal dose distribution.

As another critical innovation of the frontend VR platform, a voice assistant leverages a three-tier multimodal interface to eliminate ``controller-dependency'' during the core medical tasks.
First, a voice-driven command pipeline captures natural language queries directly through the headset’s microphone array, transmitting the raw audio to the backend for high-fidelity transcription and intent analysis.
Second, the back-end system provides bidirectional audio feedback via a low-latency TTS module, delivering real-time procedural guidance through the headset’s spatialized audio system.
Finally, the system incorporates a diegetic user interface, which projects a persistent conversation log and relevant clinical data onto 'virtual monitors' integrated into the 3D simulation environment, ensuring that critical information remains accessible without compromising the trainee’s visual presence.
% First, a voice-driven command pipeline is developed by using the VR headset's voice capability.
% When a trainee asks a question, it captures the audio and sends it to the backend for voice-to-text conversion.
% The backend provides real-time audio feedback, which the voice assistant plays to trainees.
% Additionally, the conversation log shows a diegetic UI, where response texts are projected onto "virtual monitors" within the simulation room.

\subsection{The Cognitive Backend}
% {\color{red} ---- We will rewrite this section based on the updated system design figure ----}

% The backend service serves as the ``intelligent tutor'' and uses a RAG pipeline to ground the LLM’s responses in HDR brachytherapy guidelines.
% The backend service architecture includes two core components: i) Data ingestion that constructs knowledge based on guideline documents, and ii) real-time RAG-based knowledge retrieval to provide ground truth reference for synthesizing final results.
The backend architecture functions as an ``intelligent tutor,'' leveraging a Retrieval-Augmented Generation (RAG) pipeline to anchor LLM outputs in established HDR brachytherapy guidelines. This system comprises two foundational modules: a data ingestion engine that constructs a structured knowledge base from clinical documentation, and a real-time RAG layer that provides the evidence-based references necessary to synthesize grounded final results.

% Data ingestion involves several core steps.
% First, HDR brachytherapy guidelines documents are loaded via text parsing.
% To maintain granular context, a recursive character-splitting algorithm partitions document text into 512-token chunks with strategic overlap to prevent semantic fragmentation.
% Subsequently, each chunk is mapped into a latent vector space by using a pre-trained embedding model (e.g., nomic-embed-text).
% By transforming linguistic text data into high-dimensional numerical representations, these embedding vectors encode the underlying semantic relationships of the medical protocols.
% Finally, embedding vectors are indexed in a FAISS (Facebook AI Similarity Search) vector store to enable high-speed, similarity-based retrieval.
\subsubsection{Data Ingestion}
This stage consists of several core steps.
First, HDR brachytherapy guideline documents are loaded and parsed using text processing techniques.
To maintain granular context, a recursive character-splitting algorithm partitions document text into 512-token chunks with strategic overlap to ensure semantic continuity across segments.
Subsequently, each chunk is mapped into a latent vector space by using a pre-trained embedding model (e.g., nomic-embed-text).
By transforming linguistic text data into high-dimensional numerical representations, these embedding vectors encode the underlying semantic relationships of the medical protocols.
Finally, embedding vectors are indexed in a FAISS (Facebook AI Similarity Search) vector store to enable high-speed, similarity-based data retrieval.

\subsubsection{RAG System}
The knowledge construction and data retrieval stage adopts a typical RAG pipeline.
The RAG system for knowledge retrieval contains four main stages: data preprocess, knowledge retrieval, actionable generation, and post-process, as shown in Figure \ref{fig:hdr_rag_platform}.
During preprocess stage, a high-performance STT module (e.g., OpenAI Whisper \cite{radford2023robust}) transcribes raw, unstructured audio data into normalized, structured text content.
This conversion is essential for downstream operations, enabling the RAG system to perform real-time semantic chunking and context-aware retrieval based on normalized query text.

% The RAG-based knowledge retrieval system comprises four main stages: data preprocessing, knowledge retrieval, actionable generation, and post-processing, as illustrated in Figure \ref{fig:hdr_rag_platform}. During the preprocessing stage, a speech-to-text (STT) module (e.g., OpenAI Whisper \cite{radford2023robust}) converts raw voice input from the VR user into structured textual queries. This normalization step enables downstream semantic processing and retrieval.

% The knowledge retrieval stage adopts a typical RAG pipeline.
% Upon receiving a normalized text query from the preprocess stage, the system maps the input into the same latent vector space as the knowledge base.
% Then, a Cosine Similarity search is performed to calculate the proximity between the query vectors and the indexed document vectors from the vector database.
% Finally, the Top-$k$ most relevant semantic chunks are retrieved, which provide the necessary factual grounding for the subsequent actionable generation phase.

In the knowledge retrieval stage, the RAG system used a hybrid retrieval mechanism that combines dense (embedding-based) retrieval and sparse (keyword-based) retrieval to improve accuracy and robustness.
Upon receiving a normalized text query from the preprocess stage, the system maps the input into the same latent vector space as the knowledge base.
Then, a Cosine Similarity search is performed to calculate the proximity between the query vectors and the indexed document vectors from the vector database.
Finally, the Top-$k$ most relevant semantic chunks are retrieved, which provide the necessary factual grounding for the subsequent actionable generation phase.

% At the actionable generation stage, the retrieved chunks are fed into an LLM model (e.g., GPT-4o or DeepSeek-r1), which synthesizes the retrieved context into a concise, procedurally grounded response.
% This generation process is directed by a well-defined system prompt to ensure the output is formatted as actionable instructions suitable for verbal delivery.

% Finally, post-process uses a TTS module to synthesize high-fidelity audio from generated text by LLMs, which is streamed back to the VR headset.
% This closed-loop system ensures that trainees receive immediate, evidence-based guidance while maintaining their focus on procedural workflows in the immersive HDR brachytherapy simulation.

During the actionable generation stage, the retrieved chunks are fed into an LLM model (e.g., GPT-4o-mini), which synthesizes the retrieved context into a concise, procedurally grounded response.
A well-defined system prompt directs this generation process to ensure the output is formatted as actionable instructions suitable for verbal delivery.

Finally, in the post-processing stage, a text-to-speech (TTS) module converts the generated textual response into high-quality audio output. This audio is streamed back to the VR headset, completing the interaction loop. The system enables trainees to receive immediate, context-aware, and evidence-based guidance while maintaining immersion and focus on procedural workflows.

%%------------ This section introduce following parts:
%% Prototype implementation
%% Evaluation methods and metrics: 
%% Experimental analysis based numerical results

\section{Results and Analysis}
\label{sec:experiment}
%% Development tools (Both VR and AI service)
%% Testbed setup (local GPU desktop configuration and environment setup)
%% HDR VR training platform demo case: Kepha
\subsection{Prototype Implementation and Demonstration}
\begin{table}[t] % [t] places it at the top of the page
\centering
\caption{Experimental Environment Configuration}
\vspace{-5 pt}
\label{tab:experment}
\begin{tabular}{l p{5.3cm}}
    \hline
    \textbf{Component} & \textbf{Specification} \\
    \hline
    Meta Quest 3 & RAM: 8GB; Storage: 512GB; Refresh Rate: 72Hz; Wi-Fi 6E, Bluetooth 5.2; 2 built-in speakers; and 1 built-in microphone. \\
    \hline
    WorkStation & CPU: 24 core Intel(R) Core(TM) Ultra 9 @ 6.5GHz; GPU: NVIDIA GeForce RTX 5090, 32GB; Memory: 32GB; OS: Ubuntu 22.04 \\
    \hline
    VR IDE  & Unity Hub and Meta XR Core SDK \\
    \hline
    Libraries & Langchain, PyPDF, FAISS, Ollama, Flask, OpenAI whisper, FastAPI, ngrok \\
    \hline
\end{tabular}
\vspace{-10 pt}
\end{table}

\begin{figure}[t]
\centerline{\includegraphics[width=0.48\textwidth]{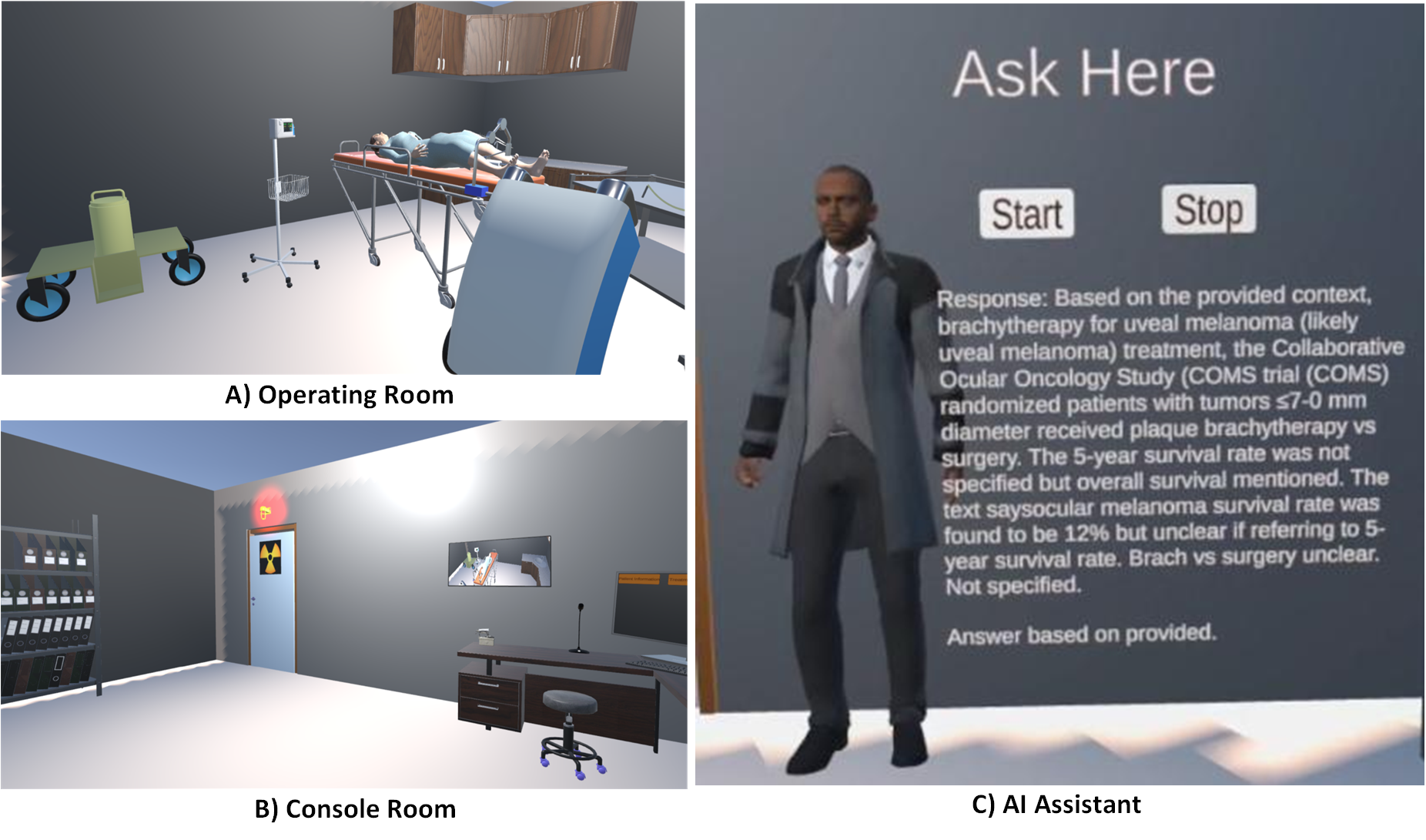}}
\caption{The HDR virtual training platform at the VR headset}
\label{fig:hdr_demo_case}
\vspace{-10 pt}
\end{figure}

The prototype was developed as a distributed system to balance the mobility of a standalone VR headset with the intensive computational requirements of the AI backend service.
We use core Meta SDKs for Unity \cite{metasdk} to develop the virtual environment and test it on Meta Quest 3 and Quest Pro.
The frontend communicates with the backend service via asynchronous RESTful APIs secured by ngrok's API gateway \cite{ngrok}.
The backend service was implemented in Python, deployed, and tested on a local workstation powered by an RTX 5090.
We use openai-whisper \cite{whisperGit} to develop STT and TTS modules.
The FAISS library \cite{douze2025faiss} is used to develop a knowledge base engine that supports similarity search and clustering of dense vector stores.
We utilize the OpenAI API architecture to access LLMs and embedding models.
The experimental setup is summarized in Table \ref{tab:experment}.

Figure \ref{fig:hdr_demo_case} demonstrates cases of the HDR virtual training platform at the VR headset.
Figure \ref{fig:hdr_demo_case}a) shows an immersive HDR VC brachytherapy operating room containing a patient stretcher and a 3D female patient model, a lead storage container and the HDR after loader.
A trainee can use touch controllers to interact with a high-detail 3D HDR afterloader and catheters on the trolley, such as plugging in or detaching applicator cables.
Figure \ref{fig:hdr_demo_case}b) shows a simulated console room with medical storage, live video monitoring setup, improved lighting, and an early-stage physical console mock-up for controlling treatment equipment.
As illustrated in Figure \ref{fig:hdr_demo_case}c), a virtual AI assistant is positioned in the control room that captures the trainee's spontaneous verbal inquiries as raw acoustic data and subsequently transmits them to the backend server.
Upon receiving a response from the backend server, it delivers high-fidelity auditory guidance and projects the transcribed text onto the screen for visual reinforcement.

The immersive training system is currently implemented using Meta's VR ecosystem as the foundation for the training platform.
which provides strong support for interactive environments, device integration, and performance optimization.
However, VR ecosystems, the system remains partially dependent on Meta’s long-term policy direction, SDK evolution, and hardware/software support lifecycle. 
To address this dependency problem, the VR system has been designed with a modular architecture, where the VR interface layer is decoupled from the core AI and backend services . This separation allows the non-VR components to remain fully portable across platforms.
We have also tested key system functionalities across different VR platforms to assess portability. These tests indicate that the core interaction and training modules can operate effectively beyond Meta-specific hardware and runtime environments with minimal modifications.
This suggests that the system is not tightly coupled to a single VR vendor and can be adapted to other VR frameworks if required.
In particular, if future changes in Meta’s policies, SDK structure, or platform support introduce constraints, the system can be transitioned to alternative VR ecosystems with limited changes to the VR interface layer.

%% This draft focuses on the two critical pillars of your evaluation: Technical Latency (crucial for VR immersion) and RAG Efficacy (crucial for medical safety).
%% Network latency: Kepha
%% Reference latency: Manoj
\subsection{System Latency and Real-Time Performance}

\begin{table}[t]
    \centering
    \caption{End-to-end System Latency (seconds).}
    \vspace{-5 pt}
    \label{tab:latency}
    \begin{tabular}{|c|c|c|c|c|c|}
        \hline
        \multicolumn{2}{|c|}{\textbf{Network Latency}}
        & \multicolumn{2}{|c|}{\textbf{Inference Latency}} & 
        \multicolumn{2}{|c|}{\textbf{Total}} \\
        \hline
        \textbf{Min} & \textbf{Max} & \textbf{Min} & \textbf{Max} & \textbf{Min} & \textbf{Max} \\
        \hline
        1  & 2 & 2 & 3 & 3 & 5 \\
        \hline
    \end{tabular}
\vspace{-10 pt}
\end{table}

We measured end-to-end latency to ensure a seamless immersive simulation experience.
The latency is the time between the completion of the user's voice/text query on the Meta Quest 3 and the delivery of the RAG-generated response.
We conduct 50 Monte Carlo runs to evaluate the average latency, as shown in Table \ref{tab:latency}.
The total latency can be divided into network latency and inference latency.
Empirical testing conducted on our local testbed, where a Meta Quest 3 interfaced with a dedicated backend server via a Local Area Network (LAN), revealed a network-related delay of 2-3 seconds.
This variability is primarily determined by: i) the duration of the captured audio payload; ii) the processing time of STT and TTS; and iii) the fluctuating network delays.
The inference latency comprises two primary computational phases: the retrieval latency incurred during semantic search across the FAISS vector index, and the generative latency associated with the LLM's contextual synthesis.
Empirical testing indicates a total temporal overhead of 1-2 seconds.
% Inference latency includes retrieval latency incurred by FAISS vector search, and storage and generation latency incurred by the LLM's response generation. The total latency is about 1-2 seconds.
Therefore, the total latency of 3-5 seconds is acceptable when ``thinking'' animations are added to mitigate an unnatural feeling in practical scenarios.
Based on domain-expert feedback, the system latency is suitable for HDR training scenarios.

% The network latency is about 3~5 seconds, depending on the length of the captured audio from the user and the network conditions. The delays are about 3~5 seconds based on our testbed, where a VR headset is connected to the local backend server via LAN networks.

%% RAG system performance evaluation: Manoj

\subsection{RAG-Enhanced Assistant Efficacy}

\begin{table}[t]
    \centering
    \caption{RAG Evaluation Dataset Categories.}
    \vspace{-5 pt}
    \label{tab:dataset}
    \begin{tabular}{|c|c|c|c|}
        \hline
        \textbf{Difficulty}  & \textbf{Basic} & \textbf{Medium} & \textbf{Advanced} \\
        \hline
        \textbf{Number}  & 31 & 11 & 10 \\
        \hline
    \end{tabular}
\vspace{-10 pt}
\end{table}

\begin{table}[t]
    \centering
    \caption{Comprehensive Metric Comparison Across Models.}
    \vspace{-5 pt}
    \label{table:rag_evaluation}
    % \begin{tabular}{|p{1.0cm}|p{1.0cm}|p{1.0cm}|p{1.0cm}|p{1.0cm}|p{1.0cm}|}
    % \begin{tabular}{|c|c|c|c|c|c|}
    %     \hline
    %     % Multirow cell in the first column, spanning 2 rows
    %     \multirow{2}{*}{\textbf{Model}} & 
    %     \multirow{2}{*}{\textbf{Embedding}} & 
    %     \multicolumn{2}{|c|}{\textbf{Retrieval}} & \multicolumn{2}{|c|}{\textbf{Generation}} \\% Multicolumn header spanning 2 columns and 3 columns
    %     \cline{3-6} % Partial horizontal line for the multicolumn area
    %     % & \textbf{Precision} & \textbf{Recall} & \textbf{Context Relevance} & \textbf{Completeness} & \textbf{Answer Accuracy} \\
    %     && \textbf{CP} & \textbf{CR} & \textbf{AR} & \textbf{AC} \\
    %     \hline
    %     % First row of the multirow data
    %     gtp-4o-mini & nomic-embed-text:v1.5 & \textbf{0.68} & 0.93 & \textbf{0.82} & \textbf{0.79} \\
    %     \hline
    %     % Another example with a different multirow
    %     gtp-3.5-turbo & nomic-embed-text:v1.5 & 0.67 & \textbf{0.90} & 0.82 & 0.74 \\
    %     \hline
    %     % First row of the multirow data
    %     gtp-4o-mini & MedEmbed-large-v0.1 & 0.62 & 0.82 & 0.83 & 0.74 \\
    %     \hline
    %     % Another example with a different multirow
    %     gtp-3.5-turbo & MedEmbed-large-v0.1 & 0.61 & 0.87 & 0.86 & 0.90 \\
    %     \cline{3-6}
    %     \hline
    \begin{tabular}{|c|c|c|c|c|}
        \hline
        \multirow{2}{*}{\textbf{Model}} & 
        \multirow{2}{*}{\textbf{Embedding}} & 
        \multicolumn{1}{|c|}{\textbf{Retrieval}} & 
        \multicolumn{2}{|c|}{\textbf{Generation}} \\
        \cline{3-5}
        && \textbf{CR} & \textbf{AR} & \textbf{AC} \\
        \hline
        
        gtp-4o-mini & nomic-embed-text:v1.5 & \textbf{0.93} & {0.82} & {0.79} \\
        \hline
        
        gtp-3.5-turbo & nomic-embed-text:v1.5 & {0.90} & 0.82 & 0.74 \\
        \hline
        
        gtp-4o-mini & MedEmbed-large-v0.1 & 0.82 & 0.83 & 0.74 \\
        \hline
        
        gtp-3.5-turbo & MedEmbed-large-v0.1 & 0.87 & \textbf{0.86} & \textbf{0.90} \\
        \hline

        % % Another example with a different multirow
        % gtp-5o-mini & 0.58 & 0.89 & 1.00 & 0.89 & 0.34 \\
        % \cline{2-6}
        % \hline
    \end{tabular}
    \begin{tablenotes}
        \small
        \item Metrics include  Context Recall (CR), Answer Relevance (AR) and Answer Completeness (AC)
    \end{tablenotes}
    \vspace{-10 pt}
\end{table}

We have 52 question-answer pairs for RAG evaluation, developed by medical professionals (domain experts) from brachytherapy training documents. These questions fall into three categories: basic, medium, and advanced levels and are intended to support medical professionals in practicing and enhancing their brachytherapy expertise. Table \ref{tab:dataset} provides a summary of the dataset.

%%% We focus on the following four metrics from the RAGAS framework \cite{Ragasdoc}.
We evaluated the performance of our RAG-enhanced assistant system by using the RAGAS framework \cite{Ragasdoc}.
We used retrieval and generation metrics to evaluate how well the system retrieves relevant context and how accurately and relevantly it generates answers based on that provided context.
% A higher value is better for all of them.
Table \ref{table:rag_evaluation} shows the performance of retrieval and generation for various LLMs and embedding models used in our system.

% \textbf{Context Precision:} It helps to determine the retriever’s ability to provide relevant contexts for response generation.
% It is calculated as a $Precision@K$ that provides out of the top $K$ items retrieved, how many of them are actually relevant to a given query, where higher scores indicate that relevant chunks are retrieved at the top position and irrelevant information is neglected. 
% Based on our observations, the highest CP of 0.68 indicates moderate precision.
% This suggests that the retriever generally retrieves useful information for answer generation, but using top-K chunks may still introduce some extra content.  

\textbf{Context Recall:}
It indicates how accurately the retriever retrieved all relevant information sufficiently to answer the query.
High recall implies that the retrieved contexts contain most of the important and necessary information needed to address the user query.
We found that recall successfully captured the required information to answer the question, with a score above 0.93 showing a strong retrieval capacity for retrieving sufficient clinical content.
Recall is a critical metric in the medical domain; missing key procedural information can lead to unsafe guidance.

\textbf{Answer Relevance:} Measures how relevant the generated response is to the user query.
A higher answer indicates better alignment with the user query, which ranges from 0 to 1.
It calculates the cosine similarity between the user query’s embedding and the embedding of the generated response.
We observed that the AR score is 0.87, which indicates that the model-generated response is closely aligned with the intent of the user’s query.

\textbf{Answer Completeness:}
It determines that the generated response covers all the different key points required to sufficiently answer the user query, as defined in the reference answer.
The reference answer is broken down into a set of key points or key information.
The generated answer is then evaluated to see whether each key point is present, directly or indirectly.
AR alone is not enough to evaluate the quality of response.
There, we evaluated answer completeness, and AC scores were consistently high across the different model settings and configurations.

We have used two different embedding models, nomic-embed-text:v1.5 is a general purpose model and MedEmbed-large-v0.1 is a domain specific embedding model trained on medical data with two large language models.The results demonstrate that nomic-embed-text:v1.5 has better retrieval performance, while MedEmbed-large-v0.1 can provide more domain-specific and concrete medical information. thus achieving higher answer completeness and relevance. Overall, the experiments show that both approaches produce high-quality answers, with domain-specific embedding increasing the quality of responses for medical-related queries.

% 1) We evaluated the RAG system's ability to incorporate HDR Brachytherapy clinical guidelines compared to a standard (non-RAG) Large Model

% 2) We will evaluate the performance of knowledge retrieval, like Accuracy \& Relevance, by comparing with ground truth Q\&A datasets.

% --- Overall LLM model performance results: Completeness, relevance, 
% --- Embedding model and Retrieval: Recall(↑)

%% Discussion: security issues and limitations: Xinyun Liu
\subsection{Limitations and Discussions}
% \Xinyun{Added by Xinyun }

% Despite the promising results demonstrated by the proposed agentic AI-driven immersive training platform, several limitations remain and deserve further discussion.
% First, the current system evaluation primarily focuses on system-level performance metrics, including end-to-end latency and RAG-based response quality, rather than clinical learning outcomes. While the reported latency and retrieval-generation metrics indicate that the platform is technically feasible for real-time interaction, we have not yet conducted large-scale user studies involving medical trainees or domain experts. 
% Second, although the RAG-based knowledge assistant significantly mitigates hallucination risks by grounding responses in authoritative clinical guidelines, its knowledge coverage is inherently constrained by the curated document corpus. The current implementation relies on standardized HDR brachytherapy guidelines, which may not fully capture institution-specific protocols, practitioner preferences, or rare edge cases encountered in real-world clinical practice.
% Third, the current prototype emphasizes procedural guidance and knowledge retrieval, but does not yet incorporate automated assessment or adaptive feedback mechanisms. While trainees can receive real-time instructions, the system does not quantitatively evaluate procedural correctness, spatial precision, or radiation safety compliance during simulation. 
% Future work will involve controlled user studies and expert assessments to evaluate educational effectiveness and clinical realism more rigorously.

Despite the promising results demonstrated by the proposed agentic AI-driven immersive training platform, several limitations remain and deserve further discussion.
First, the current system evaluation primarily focuses on objective metrics, such as end-to-end latency and AI assistant response quality. While these metrics provide useful insights into technical performance, end-to-end system validation has so far been conducted through a limited user study involving a domain expert (physician).
This is the expert-level feedback provided by the initial evaluation. It does not yet capture learning effectiveness and variability across user groups. To remedy this deficiency, we intend to gather data through an expanded user study including several hundred medical trainees in future work. This will allow for a more comprehensive assessment of educational impact on learner, user experience, and generalizability across different levels of clinical expertise.

Second, the RAG-based knowledge assistant minimizes the risk of hallucinations by grounding its responses in authoritative clinical guidelines.
However, its ability to provide answers is ultimately limited by the quality and number of documents it can reference.
As it is based on a standardized set of HDR brachytherapy guidelines, it cannot account for institution-specific protocols, practitioner preferences, or unusual scenarios that occur in clinical practice.
Future work will integrate multi-institutional datasets and practitioner feedback to reconcile standardized guidelines with localized protocols and rare clinical complexities.

Third, current VR technology is incapable of tracking users' hand movements in virtual real-world environments.
It emphasizes providing users with procedural guidance to assist users in recalling information; however, there is still no automated assessment/assessment or adaptive feedback mechanism.
Although users may receive real-time instructions as they perform procedures, the current system does not quantitatively evaluate procedural correctness and spatial precision.
To address these constraints, future efforts will focus on integrating advanced hand-tracking sensors and AI-driven kinematic analysis to enable automated, quantitative assessment of procedural accuracy and spatial precision.

\section{Conclusions and Future Work}
\label{sec:conclusion}
This paper presents an immersive and knowledge-aware stimulation platform to enhance the pedagogical efficacy of HDR brachytherapy education.
A hands-free, interactive environment running on a VR headset allows trainees to navigate procedural training workflows and access clinical guidelines via natural language.
While an RAG-enhanced AI assistant provides grounded and relevant responses, effectively mitigating the risks of LLM hallucinations under the HDR brachytherapy context.
The experimental results based on a proof-of-concept prototype demonstrate acceptable system latency and high context precision and answer relevance.

On reporting the current version of prototype implementation, our ongoing efforts focus on several directions.
First of all, we will conduct controlled user studies to compare the outputs of the proposed solution with those of traditional HDR brachytherapy training methods and evaluate specific educational metrics.
We will also engage board-certified professionals to assess the ``clinical realism'' of the AI's guidance.
To address the ``knowledge coverage'' constraint, we will develop hybrid knowledge retrieval models that enable the agent to access the latest external literature for rare edge cases not covered by the primary guidelines.
Additionally, we will introduce Expert-in-the-Loop (EITL), where experts can evaluate AI responses and improve the agent's accuracy through Reinforcement Learning from Human Feedback (RLHF).
Last but not least, we will develop VR modules for additional HDR applications requiring different procedures and techniques.

% \section*{Acknowledgment}

% The preferred spelling of the word ``acknowledgment'' in America is without 
% an ``e'' after the ``g''. Avoid the stilted expression ``one of us (R. B. 
% G.) thanks $\ldots$''. Instead, try ``R. B. G. thanks$\ldots$''. Put sponsor 
% acknowledgments in the unnumbered footnote on the first page.
\bibliographystyle{IEEEtranS}
\bibliography{02_references,02_VR_training}

% \vspace{12pt}
% \color{red}
% IEEE conference templates contain guidance text for composing and formatting conference papers. Please ensure that all template text is removed from your conference paper prior to submission to the conference. Failure to remove the template text from your paper may result in your paper not being published.

\end{document}